# Extending the Abstraction of Personality Types based on MBTI with Machine Learning & Natural Language Processing (NLP)

**Carlos Basto**

## Abstract

This paper presents a data-centric approach with Natural Language Processing (NLP) to predict personality types based on the MBTI (an introspective self-assessment questionnaire that indicates different psychological preferences about how people perceive the world and make decisions) through systematic enrichment of text representation, based on the domain of the area, under the generation of features based on three types of analysis: sentimental, grammatical and aspects. The experimentation had a robust baseline of stacked models, with premature optimization of hyperparameters through grid search, with gradual feedback, for each of the four classifiers (dichotomies) of MBTI. The results showed that attention to the data iteration loop focused on quality, explanatory power and representativeness for the abstraction of more relevant/important resources for the studied phenomenon made it possible to improve the evaluation metrics results more quickly and less costly than complex models such as the LSTM or state of the art ones as BERT, as well as the importance of these results by comparisons made from various perspectives. In addition, the study demonstrated a broad spectrum for the evolution and deepening of the task and possible approaches for a greater extension of the abstraction of personality types.

## 1. Introduction

In this section, a description of the field of study, hypotheses and their dependencies/links are presented, in addition to their main motivations. Furthermore, the strategy for addressing the hypotheses is devised and findings, challenges and details will also be provided.

### 1.1. Psychological Types

This study focuses on the essential basis for the classification of psychological types – that is, the basis for determining a broader spectrum of characteristics derived from a human being (Jung and Baynes, 1953)– through the use of machine learning and natural language processing. This topic is crucial in differential psychology and has been extensively worked on by Carl Jung (further details and correlated studies available no IAAP[1]) which allows us to ensure that the psychological classification of different types of individuals is a promising step towards self-knowledge.

Under Jung's proposal (Geyer, 2014), a dominant function, along with the dominant attitude, characterizes consciousness, while its opposite is repressed and characterizes the unconscious. This dualism is extremely valuable in understanding and, eventually, predicting personality types. Thus, this classification opened doors for further enrichment of his work.

In the 1920s, Katharine Cook Briggs, an American writer interested in human personality, developed her own theory of personality types before learning about Jung's writings. Then, together with her daughter, Isabel Briggs Myers, they developed a convenient way to describe the order of each person's Jungian preferences (Geyer, 1995) and thus the four-letter acronym was born.

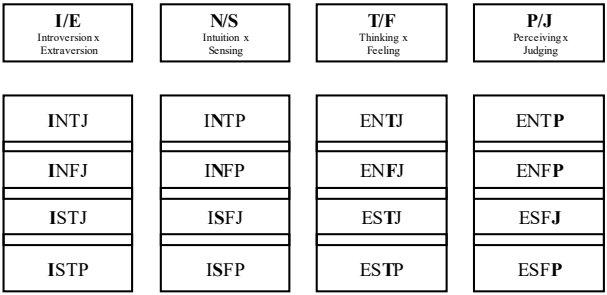

Fig. 1. Four central dichotomies and their 16 possible combinations for determining the types of personalities

The phenomenon studied is extremely engaging and comprehensive and portrays part of the human complexity for self-knowledge, therefore, the following experiments were focused on the area/domain and on the specific and recursive attention of the representativeness of the variables for, through the application of artificial intelligence resources, converge the part of human intelligence (or demonstration of it) with the intelligent machine learning resources in a relevant and robust way.

---

[1] International Association for Analytical Psychology (IAAP)

## 1.2. Related Work

According to the Oxford English Dictionary (OED)[2], personality consists of "the various aspects of a person's character that combine to make them different from other people". In other words, personality is the expression of the individual. Therefore, the problem defined as the scope of this study is comprehensive and audacious, since the determination of a broader spectrum of derivation of the characteristics of human beings brings to light a series of possibilities for action and also approaches for its understanding. Therefore, it was necessary to validate, previously, its relevance from the perspective of psychology. Yes, there are countless studies, articles and other scientific research that promote work on this human aspect; however, as part of the responsibility of this study, prior validation offered an extension of the meaning of the techniques proposed for the research and ratified the tasks that are provided to support the study, in addition to increasing the knowledge obtained with the project.

The differentiation between inference and traditional assessments or methods for predicting personality types ([Novikov et al., 2021](#)) was assessed and, as a result, it was identified some points of attention: (i) distance from the prediction and from the real personality traits (error) (ii) lack of previous hypotheses, (iii) retention of observed patterns for generalization, (iv) low success and progress of the machine learning approach for personality prediction and (v) high success discrepancy between training, validation and underfitting.

The observed literature corroborated by means of two main biases: the psychological and the technical (machine learning - or intelligent systems). From the psychological point of view it helped to, predominantly, compose the grounded basis of this study under a comparative approach of traditional methods of evaluating the types of human personality with the application of automated methods as well as the description of use innovative and relevant to the approach such as ([Dhelim et al., 2021](#)), ([Shuster et al., 2019](#)) or ([Lima and Castro, 2021](#)). Entering the field of Artificial Intelligence and its resources, the research brought to light the feature selection and models and, even, the methods of self-attention based feature selection together with a deeper view on the selection of discriminative and generative models. Such work did not justify or prefer the use of one or the other approach in relation to the field of study addressed; therefore, the literature, in general, contributed to the deepening of possible model selection techniques, in addition to the optimization of hyperparameters and their feasibility in the world of psychology.

Related work has shown that the difference in performance is not usually significant, that the quality of the data is rarely aimed at as the focal point of the experiment and that, in general, the metrics for evaluating the models tend to remain within the expected standards and consistencies for the classification problem.

## 1.3. Hypotheses Genesis

Of all the challenges observed and brought to this study, there is clearly a strong work in data manipulation and techniques to generate or use text resources for textual analysis. This type of work offers more orthodox ways of using syntactic ([Liesting et al., 2021](#)) information: study of dependency trees, textual correspondences, morphology, sentence structure, among other various techniques and study approaches for example. However, new approaches to abstraction of the entities found in the text seem to be emerging with increasing strength.

In addition to pattern abstraction and resource engineering, there is a very expressive focus on improving models (performance, accuracy). Examples of these initiatives are the use of Pre-trained Models (PTMs) ([Dai et al, 2021](#)) and ([Keh et al., 2019](#)), hybrid scenarios and/or mixed generative and discriminative models ([Mesnil et al., 2015](#)). However, regardless of the approach used, they all show significantly positive and reasonable results. Many assessments were based on accuracy, which, in unbalanced scenarios, does not guarantee the robustness of the model for later generalization.

This may suggest a space or, perhaps, even a request to focus on working with the data (with the implementation of more data-centric techniques) that serve good resources to the models and not just the models themselves. Thus, it is possible to explore deeper or more specific or even unknown polarities. Point that the performance improvement to the model does not promote.

This contradiction, unfortunately, portrays the superficiality of the approach to the subject justified by its bases of reference (such as classification techniques) and taking them as absolute or partially absolute. Also with traditional methods, problems come up with different classifications depending on the techniques used or the difficulty of detecting moods even considering the advances in the area (Akula and Garibay, 2021). Most of the available estimates refer to the degree of agreement of the expected characteristics with the gold standard labels – a dictionary that guarantees a reasonable basis for searching for personality in texts originated from the conventional use of self-assessment tests or external annotators.

---

[2] Oxford Learner's Dictionaries (2021) [Personality](#).

Personality inference based on digital data can be an alternative to conventional psychological tests, but it is not complete or absolute. So, designing a good dictionary and data may accelerate the focus on what matters and, after some research, it was possible to detect the most important aspects of quality conversational, namely: (i) repetition, (ii) interest, (iii) making sense, (iv) fluency, (v) listening (See et al., 2019). These characteristics of a good conversation portray, in part, what we need to have human success in a conversation. And taking this as a basis for generating features can mean better abstraction and, therefore, better results in detecting personality patterns.

The research question to be answered by this study was based on what is already known about data-centric models and their benefits in improving the model's results for predicting personality types and text classification.

From the broad study carried out, it was possible to establish 2 (two) hypotheses: one conceptual and one technical. The conceptual one brings the idea that a phenomenon such as difficult and complex personality types can effectively have better results and significantly approach the real with robustness (generalization power):

> "Handcrafted feature representations have positive effect on personality type prediction"

Fig. 2. Domain-based (or conceptual) hypothesis

The above hypothesis is justified by the context of this type of prediction studied in the literature review - personality types. Thus, it was possible to conceive that machine learning models tend to make many mistakes, although their technical evaluations are relatively good. One of the possible motivations for this is that, since the model generalizes new data, it is very difficult to "check" whether the model was right or not. For this reason, many models are completed overestimated, even if they are poorly adjusted, the robustness of their predictions cannot be measured so simply.

The idea of generating handcrafted features concerns the field of study and its abstraction. In order to be able to positively improve the result of predictive models in the field of psychology and its extension, it is necessary to perform the best possible abstraction of the phenomenon through the data so that the pattern identified is indeed robust for application in new data. In NLP, pre-processing and text representation tasks do not fully meet this requirement; so that the generation of deep features that portray the perspectives of greater impact or influence to the observed phenomenon fulfills this gap. Additionally, a special attention to the data that the dataset was sampled goes against the technical hypothesis:

> "Data-centric models lead to faster improvements on predictive models"

Fig. 3. Technical hypothesis

Currently, it is observed that many initiatives in favor of data exist and are growing. However, this happens in many other areas, such as Business Intelligence for example. Data is vitally important[3], although many cares more about what to do with it, rather than worrying about it itself.

## 1.4. Applied Approach

A feature that expresses, with a high level of precision, a behavior of the phenomenon we are studying has a greater weight (or more positive effect) than thousands that represent a little of what has been studied. Obviously, these claims may seem extreme and perhaps utopian; but, through an artery of the research core of this study, the next pages will show a little of the tests and achievements that have been obtained in this regard.

The studied literature was able to provide the necessary resources to understand, from psychology perspective, what matters most and thus establish the essential points for the good result of the project: (i) model, (ii) data and (iii) dictionary.

Even though a dictionary[4] is part of the data, a different treatment is expected due to its high importance and impact on the result of the application of intelligent systems in the prediction of personalities. Also, the segregation of data and dictionaries portrays the difference in their generation and distribution, in addition to delimiting a line of work and focus for the sake of data orientation.

---

[3] Data is the new oil may give further motivations for that.

[4] In other words, dictionary is the method which classification is possible. For this study, MBTI was used.

Once dictionary is already established, focusing on good data is the next step. This work can be performed more focused on what matters, but what to expect from them and what to look for? For that, a Natural Language Generation (NLG) point of view can help (through a semantic reverse engineering) to obtain the most important aspects of conversational quality related to the project: (i) repetition, (ii) interestingness, (iii) sense, (iv) fluency and (v) listening. This approach was used as a benchmark for how human expression occurs optimally during a conversation (albeit with a different application). These characteristics of a good conversation portray, in part, what we need to humanly be successful in a conversation. And taking this as a basis for the generation of features can mean better abstraction and, therefore, better results in personality patterns detection. Under these variables, it was possible to conceive the steps used to address the hypotheses:

**Algorithm 1** Strategy for Data Iteration Loop

Description of the approach to abstract features from the knowledge of psychology focused on the extension of this abstraction. In other words, the strategy used for designing, along the experiments, the text representation (how are the steps performed on data to achieve a good text representation for the study).

**DATA CLEANING**: for handling **repetitions**, punctuation, stop words, URLs etc.

**PRE-PROCESSING:** work on word treatment, topic modeling and/or lemmatization for dealing with **fluency** and **listening**.

**FOR** each experiment **DO:**

> **WEIGHTED DICTIONARY:** generate a dictionary of words, assign meanings and from well-defined aspects and grammar practices and techniques, project levels and classes of **sense**.
>
> **DOMAIN ASSESMENT:** generate a dictionary of words, assign weights based on well-defined psychological practices and techniques, to design levels and classes of **interestingness**.

**ENDFOR**

Imagine that you do not have such a wide vocabulary and want to refer to a briefcase, but you only know "suitcase" or "box". Or, in your language, you don't have words alone that represent what you need. This will have an impact on the "interpretation" of the evaluated text. So the richer your dictionary is the most promising it can be in the classification task. Well, the algorithm is proposed to (or try to) avoid this lack of representation resources (bold words in Algorithm 1).

The figure 4 shows that from a loop of data improvement iterations, it was possible to abstract different values from a prism oriented to the studied domain. Keeping the model (stacked models) and working on the quality of the data, it was possible to enrich the data, initially, from its cleaning and pre-processing steps until deeper abstractions of handcrafted features that could establish the link of the best communication practices described and text representation. For each experiment an approach was applied (sentimental analysis, sentimental analysis based on aspects and grammatical analysis).

These analyzes provided new features that raised the level of quality and expressiveness of the data by improving performance to the model as a whole. A fourth experiment was conducted with the use of all features in order to generalize these specific qualities. The results were described in section "4. Results ".

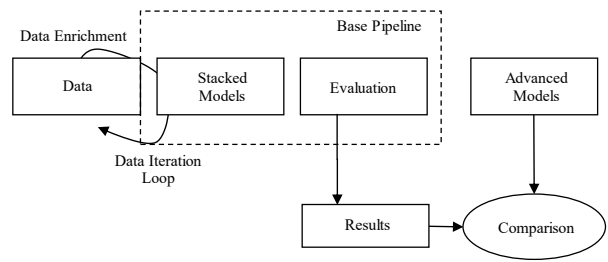

Fig. 4. Graphical visualization of the data iteration loop strategy

## 1.5. Preface

The next sections aim to cover the entire trajectory relatively based on the CRISP-DM (Chapman et al., 2000) necessary to understand the data and design the content, followed by the description of the baseline of models and related strategies for hyperparameter tuning and ensemble methods. In addition, the reference models will be cited.

In the section "4. Results ", the metrics used to evaluate the performance of the experiments will be described and positioned in accordance with the respective objectives proposed in conjunction with the ablative and error analysis strategy that were performed during the study.

The entire flow for the features abstraction was established to support study's justifications for success, containing details of sentimental, grammatical and aspects analysis as well as the union of all for conclusive purposes of evaluation results. Finally, still in this section, the comparative results are presented under feedback and points of view from a psychological and technical perspective. Finally, the subsequent pages will portray the significance, the implications, the limitations of the study and possible directions for its extension.

## 2. Data

The dataset MBTI Myers-Briggs Personality Type[5] was collected through the Personalite Café Forum[6] which contains 8675 observations and two (2) features: "type" that brings the 4 letter MBTI code/type and "posts" comprised of the last 50 things people have posted. No nulls or duplicated rows were presented in the data. Also, under total size in memory of 135.7 KiB, dataset has average record size in memory of 16 B. In the following sections both features will be described.

### 2.1. Personality Types

This variable has 16 possible values (or categories) as showed in fig 1. It was possible to detect that INFP personality type has stronger representation in dataset and data is unbalanced as it comprises 21.1% of all the data set, followed by INFJ with 16.9%, INTP with 15% and INTJ with 12.6%. These are the most representative personality types in the data set.

Breaking the classes down, it was clear most of occurring character happens to be "N" (or Intuition). It is very interesting because intuitive individuals are very imaginative, open-minded and curious. They prefer newness over security and focus on hidden meanings and future possibilities what would make it up in the occurrence list while "S" (or Sensitive) are the ones highly practical, pragmatic and down-to-earth. They tend to have strong habits and focus on what is happening or has already happened which might make it down in the list.

### 2.2. Posts

"posts" features has (as expected) a high cardinality with 8675 distinct values and it is uniformly distributed. Some "feelings" like emoticons and its frequency were evaluated and it was possible to see that "happy" ones are more frequent (such as ☺ with 15.6% frequency, for example) and money-related symbols like $ occurs 90.5% related to other ones (€, for example).

### 2.3. Holistic View

According to personality types distribution was possible to confirm that the dataset is unbalanced.

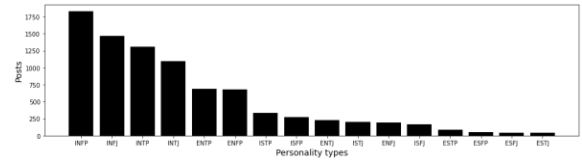

Fig. 5. Personality types distribution

Also, checking the occurrences, the distribution between personality type pairs is pretty different. Still, it is clear most posts were written between 6000-10000 characters which seemed sufficient for detecting patterns.

### 2.4. Length Analysis

Length analysis could help in detecting simple textual patterns and information about representativeness. For the dataset used, the number of words and its variance per post is not significantly different. Both occurrence and variance of words across personality types could not give any clear differences between them, suggesting words per comments are pretty much similar for all possible categories just like their variance.

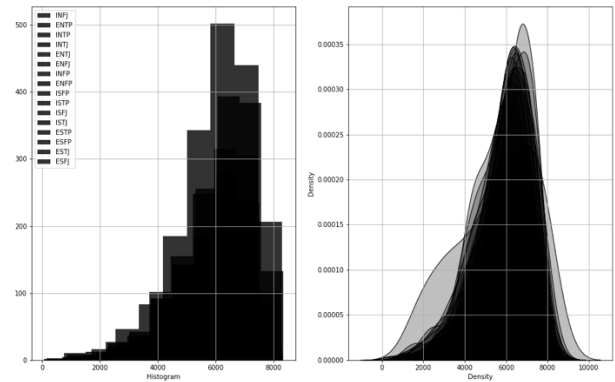

Fig. 6. Histogram and density of personality types distribution

Distribution of the variables with respect to the target are analyzed by comparing the histograms and densities of the samples. The 16 categories have a similar length distribution and density ends up showing sample different sizes.

---

[5] (MBTI) Myers-Briggs Personality Type which ncludes a large number of people's MBTI type and content written by them.

[6] PersonalityCafe Forum a forum community dedicated to all ranges of personality types and people.

## 2.5. Correlation

Finally, for this Initial Data Analysis (IDA)[7], correlation was evaluated. Pearson's correlation coefficient (r) suggested that the linear correlation between personality types represented good insights to work with them. For example, I/E people are negatively correlated with N/S and it may be due to the higher volume of extroverted (E) in the I/E group and sensing in the (N/S) group. This makes sense once the naturalness and easy receptivity to excitations of the E does not match the "down-to-earth" of an S.

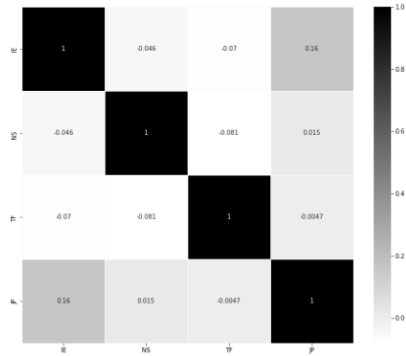

Fig. 7. Features Pearson's correlation coefficients

Another example may be TF and NS negative correlation that might suggest that the way they see the world (N/S) does not determine how they make decisions or deal with emotions (T/F).

## 3. Models

The models used for the baseline brought an initial and robust perspective of the modeling phase and served as a reference for all subsequent experiments. For the execution of the experiments, a virtual machine configured with NVIDIA-SMI 460.73.01, driver version 460.73.01, CUDA version 11.2 and 12206MiB of memory was used.

### 3.1. Baseline Composition

In this section, the text classification task will be obtained and, together with the presentation of the chosen data set, they will justify the baseline and general reasoning for the experiments conducted in addition to the introduction to the models and pipelines used.

The choice of models that make up the baseline was made based on their specificities, assumptions and frequency of use obtained in benchmarks and with the studied literature. The composition has models under a discriminative and generative approach, machine learning and deep learning, ensemble methods (bagging). This use was aimed at guaranteeing robustness and provision of resources for good prediction from the beginning. Still, more complex methods of ensemble (stacking) were used to guarantee appropriate model selection for the experiments.

The basic and intermediate validation steps to improve the performance of the models were already implemented in the base pipeline so that they did not become a relevant influence throughout the experiments since the focus was to keep the models fixed and iterate with the data (data-centric approach). Below, the models and their settings are described.

**Naive Bayes**

Naive Bayes algorithm for multinomially distributed data was considered suitable enough variant of NB used in text classification (where the data will be represented as tf-idf). The distribution was parametrized by vectors $\theta_y = (\theta_{y1}, \dots, \theta_{yn})$ for each class y, where n was set as the size of the vocabulary and $\theta_{yi}$ as the probability $P(x_i \mid y)$ of feature $i$ appearing in a sample that belongs to class y. $\theta_y$ was estimated by a smoothed version of maximum likelihood, relative frequency counting:

$$\theta_{yi} = \frac{N_{yi} + \alpha}{N_y + \alpha n}$$

where $N_{yi} = \sum_{x \in T} x_i$ was the number of times feature $i$ appeared in a sample of class y in the training set T, and $N_y = \sum_{i=1}^{n} N_{yi}$ was the total count of all features for class y. For features are not present in the learning samples and for zero probabilities prevention, model was designed through Laplace smoothing implementation. Known that the severe assumptions of the model adversely affect the quality of its results, some treatments were foreseen such as the correction of the unbalanced data, weight magnitude errors checks and terms frequency handling (Rennie et al., 2003).

**Random Forest**

This model was used considering a minimum of 100 estimators and gini (or eventually entropy) was the measure for split quality. Two sources of randomness for decrease the variance of the forest estimator and its combination by averaging their probabilistic prediction were used instead of letting each classifier vote for a single class.

---
[7] A systematic approach to initial data analysis is good research practice

$$\text{Gini} = 1 - \sum_{i=1}^{n} p^2(C_i)$$

$$\text{Entropy} = \sum_{i=1}^{n} -p(C_i) \log_2(p(C_i))$$

Where $p(C_i)$ is the probability of class $C_i$ in a node.

The maximum depth of the tree was automatically expanded until all leaves were pure or until all leaves and the maximum features threshold hit $\sqrt{n\_features}$, considering its size as $O(M * N * \log(N))$ where $M$ was set as the number of trees and $N$ as the number of samples.

**Logistic Regression**

Logistic regression model was solved with L-BFGS-B (Keskar and Waechter, 2016) and (Schmidt et al., 2017) (which is a limited-memory quasi-Newton code for bound-constrained optimization) and Newton conjugate gradient (Forsgren and Odland, 2015).

$$h_\theta(x) = \frac{1}{1+e^{-\theta^T x}}$$

For optimization problem, ℓ2 regularization was used to minimize the following cost function:

$$\min_{w,c} \frac{1}{2} w^T w + C \sum_{i=1}^{n} \log(\exp(-y_i(X_i^T w + c)) + 1)$$

And elastic net with the following one:

$$\min_{w,c} \frac{1-\rho}{2} w^T w + \rho \|w\|_1 + C \sum_{i=1}^{n} \log(\exp(-y_i(X_i^T w + c)) + 1)$$

All features needed to be incorporated in the model according to domain knowledge and that's the why the ones with minor contribution have their coefficients close to zero, expecting to perform better.

C hyperparameter (like in support vector machines, smaller values specify stronger regularization) was used ranging from 1 to 10 where $\rho$ controlled the strength of ℓ1 regularization (when elastic net was used) vs. ℓ2 regularization.

**K-Nearest Neighbors**

As non-parametric approach, KNN was used by setting it up under power parameter for the Minkowski metric $\sum (|x-y|^p)^{(\frac{1}{p})}$ where $p$ is 2 which is equivalent to the Euclidean distance $\sum ((x-y)^2)$. The weight points were measured by the inverse of their distance (closer neighbors of a query point had greater influence than neighbors which were further away). The algorithms used to compute the nearest neighbors varied among BallTree (Omohundro, 1989), KDTree or brute-force (Bentley, 1975) search and they were chosen by attempting to decide the most appropriate one based on the data during fitting process.

**Support Vector Machine**

Due to its capacity of work effectively in high dimensional spaces and its versatility (such as kernel functions) Support Vector Classifier was used.

$$\min_{w,b,\zeta} \frac{1}{2} w^T w + C \sum_{i=1}^{n} \zeta_i$$
$$\text{subject to} \quad y_i(w^T \phi(x_i) + b) \geq 1 - \zeta_i,$$
$$\zeta_i \geq 0, i = 1, \ldots, n$$

With Radial Basis Function (Hemsi et al., 2014) as kernel:

$$K(x, x') = \exp\left(-\frac{\|x - x'\|^2}{2\sigma^2}\right)$$

where γ is specified by parameter gamma set as $\frac{1}{n\_features * x\sigma}$ and hyperparameter C will be set as 1.

**Multi-Layer Perceptron**

MLP was used in contrast to logistic regression because of their main difference – between the input and the output layer, it was used non-linear layers. MLP is a relatively simple form of neural network because it is front propagation[8] only.

This model had 'relu'[9] as activation function, the rectified linear unit function which returns $f(x) = \begin{cases} 0 \text{ for } x < 0 \\ x, \ x \geq 0 \end{cases}$ and solved by 'adam' which refers to a stochastic gradient-based optimizer proposed by Kingma, Diederik, and Jimmy Ba (Kingma and Ba, 2014).

---

[8] Feedforward neural network is an artificial neural network wherein connections between the nodes do not form a cycle.

[9] Rectifier (neural networks)

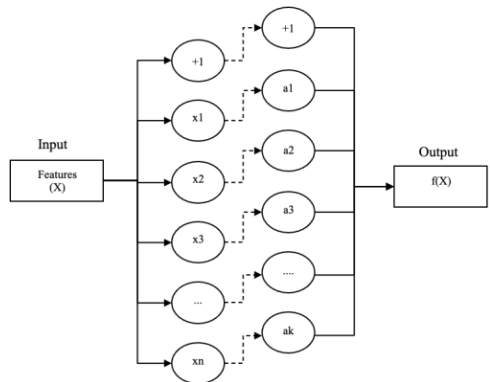

Fig. 8 Multi-Layer Perceptron architecture

Hidden layers were worked ranging between 50 and 150 with max iteration ranging from 500 to 1500.

### 3.2. Hyperparameters Tuning

Considering several methods (Anonymous, 2019), (Zhang et al., 2017) of finding the best hyperparameters for the models, it was concluded that for the assignment of hyperparameters, a search of the hyperparameter space for the best score in the cross-validation had to be conducted before any evolution in the use of models since the optimization of hyperparameters should be considered customary, as the data should drive (i) the space hyperparameter design, (ii) the method to search or sample candidates, (iii) the cross-validation scheme and (iv) the scoring function (Zahedi et al., 2021).

The technique exhaustively considered all combinations of hyperparameters from a reasonable range obtained from past experiences and benchmarks. It is common for a small subset of these hyperparameters to have a major impact on the predictive or computational performance of the model, while others may be left at their default values (Cawley and Talbot, 2010) and it was what justified pre-tunning hyperparameters strategy for modeling.

### 3.3. Ensemble Methods

After some benchmark ideas (Bouchard and Celeux, 2006) and (Ramezani et al., 2021), as it will be mentioned in the next sections, many models were underfitted, suggesting high bias was present in the data, so, stacked generalization was used due to its approach for reducing it. More precisely, the predictions of each individual estimator were stacked together and used as input to a final estimator to compute the prediction (Chatzimparmpas et al., 2020). This final estimator was trained through 5-fold cross-validation.

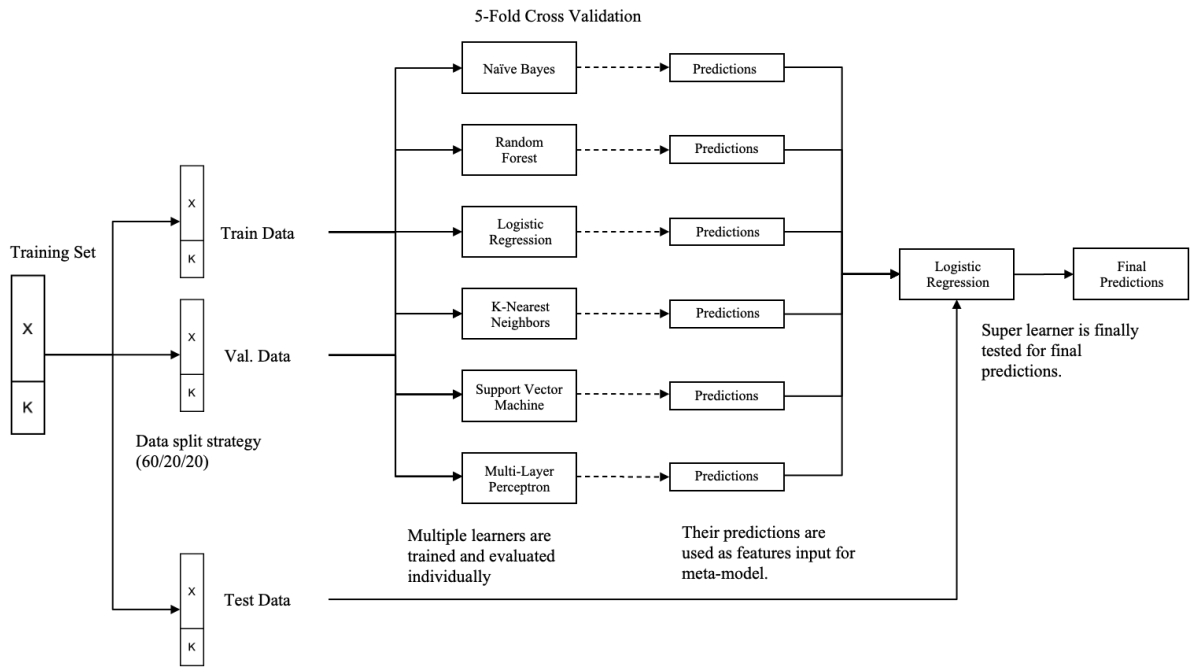

Fig. 9. Stacking architecture design - during training, the estimators are fitted on the whole training data so they can be used for predicting. To generalize and avoid overfitting, the final estimator is trained on out-samples using cross-validation.

## 3.4. Reference Models

The following models were used for reasons of comparisons exclusively with the data-centric approach and served as a reference for the diagnosis of results.

**Sequential Long short-term memory (LSTM)**

Long short-term memory (LSTM)[10] is an artificial recurrent neural network (RNN) architecture which was used as a representation of complex models (deep learning) for comparison to baseline and experiments results.

For Tokenization, a class from Keras[11] which allows to vectorize a text corpus, by turning each text into either a sequence of integers (each integer being the index of a token in a dictionary) or into a vector where the coefficient for each token based on TF-IDF was used. The maximum number of words to keep, based on word frequency was set up with 10.000. A specific string was assigned to word index and used to replace out-of-vocabulary words during text to sequence calls.

The Sequential Neural Network Architecture was designed based on ADAM optimizer with learning rate 0.00001, having embedding layer as input and followed by two LSTM layers (first with 200 nodes and second with 20), having a dense 'relu' layer with 64 nodes and finally a sigmoid layer for each of the models and binary cross-entropy as loss function. Additionally, a dropout of 0.3 was used throughout the layers.

**Bidirectional Encoder Representations from Transformers (BERT)**

To represent the state of art model for text classification in this study, one of the most used which is BERT was implemented in its base and uncased form. Thus, it used a vocabulary of around 30.000 words. The processes of tokenization (Sennrich et al.,2016) involved splitting the input text into a list of tokens that were available in the vocabulary.

An out of vocabulary word was progressively split into subwords and the word was represented by a group of subwords. Since the subwords were part of the vocabulary, it learned a context for these subwords and the context of the word which is simply the combination of the context of the subwords.

Model was set up with 12-layer, 768-hidden, 12-heads and around 110M parameters.

## 4. Results

In this section, the text classification task will be obtained and, together with the presentation of the chosen data set, they will justify the baseline and reasoning line for the experiments conducted in addition to the introduction to the models and pipelines used.

### 4.1. Metrics

Considering some preliminary tests and according to literature review, it was assumed that psychological tasks are very complex and hard to have their patterns detected. Still, high errors are always present in evaluation results for this type of initiative. The strategy adopted was to focus on classification groups (I/E, N/S etc.). In this way, it is possible to, individually, abstract personality types and work on their peculiarities. In other words, four (4) classifiers were modeled.

The evaluation of the metrics was arranged by way of exemplifying the positive impact the performance of the models from the enrichment of the quality of the data from several perspectives had on general results, namely:

**Benchmark**: many related works present their work / conclusions based on this metric. Although, for the present study, it is not safe and was maintained only for further comparisons.

$$\text{accuracy}(y, \hat{y}) = \frac{1}{n_{\text{samples}}} \sum_{i=0}^{n_{\text{samples}}-1} 1(\hat{y}_i = y_i)$$

**Model Quality:** F1-Score and its variations: the harmonic average between precision and recall portrays the general quality of the models and works well even with data sets that have disproportionate classes.

$$F1 = 2 * \left(\frac{precision * recall}{precision + recall}\right)$$

And its derivations used: (i) "micro" which was used for calculating metrics globally by counting the total true positives, false negatives and false positives, (ii) "macro" which calculated metrics for each label, and found their unweighted mean (it did not take label imbalance into account) and (iii) "weighted" which contributed with metrics for each label, and had their average weighted by support (the number of true instances for each label) – in fact, this improved "macro" to compensate for the label imbalance and resulted in an F1 score that was not exactly between accuracy and recall.

---

[10] Long short-term memory wikipedia

[11] Introduction to Keras for Researchers

**Predictive Power:** Precision aims to show how good the model was in predicting the target class and how accurate were the shots the model gives in this regard. In other words, it reflected the predictive perspective directly.

$$\text{Precision} = \left(\frac{TP}{TP + FP}\right)$$

**Predictive Power Quality:** Recall, on the other hand, showed how many shots models hit, that is, it represented the quality of the model's predictive power.

$$\text{Recall} = \left(\frac{TP}{TP + FN}\right)$$

**Approximation/Similarity:** Jaccard was used to compare the predicted data set for the target class in correspondence with what was actually 1.

$$J(y_i, y_i) = \frac{|y_i \cap y_i|}{|y_i \cup y_i|}.$$

**Distinguishing Capability:** the Receiver Operator Characteristic (ROC) curve is an evaluation metric for binary classification problems. It is a probability curve that plots the TPR against FPR at various threshold values and essentially separates the 'signal' from the 'noise'. In addition to Recall (or TPR), ROC uses FPR as follows:

$$\text{FPR} = \left(\frac{FP}{FP + TN}\right)$$

An ROC curve plots TPR vs. FPR at different classification thresholds. Lowering the classification threshold classifies more items as positive, thus increasing both False Positives and True Positives. And, finally, AUC stands for "Area under the ROC Curve." That is, AUC measures the entire two-dimensional area underneath the entire ROC curve (integral calculus) from (0,0) to (1,1).

The mixture of the two was used to represent models' distinguishing capability.

**Error Rate:** Log Loss function computes log loss given a list of ground-truth labels and a probability matrix, as returned by an estimator's prediction probability method. With the true label $y \in \{0,1\}$ and a probability estimate $p = \Pr(y = 1)$, the log loss per sample was the negative log-likelihood of the classifier given the true label and it served for evaluating errors in general.

$$\begin{aligned} L_{\log}(y, p) &= -\log \Pr(y|p) \\ &= -(y\log(p) + (1 - y)\log(1 - p)) \end{aligned}$$

Additionally, Confusion Matrix was used as holistic point-of-view on mistakes and successes that the models had and Learning Curves helped on diagnosing bias-variance.

### 4.2. Baseline Results

From a quick and more enlightening approach on what the data requires for the prediction of personality types, an initial baseline was created based on the models mentioned in "3.1. Baseline Composition" and with the first results, further data-driven tasks were performed.

*Table 1*. Results obtained in initial run which comprised the baseline for further experiments.

| Metrics (%) | IE | NS | FT | JP |
|---|---|---|---|---|
| Accuracy | 76,37 | 85,16 | 74,86 | 63,69 |
| F1-Score Macro | 49,32 | 46,94 | 74,51 | 57,16 |
| F1-Score Micro | 76,37 | 85,16 | 74,86 | 63,69 |
| F1-Score Weighted | 68,85 | 78,47 | 74,80 | 60,75 |
| F1-Score | 12,00 | 2,00 | 72,00 | 74,00 |
| Jaccard Score | 6,55 | 0,96 | 55,71 | 58,59 |
| Log Loss | 8,16 | 5,13 | 8,68 | 12,54 |
| Precision | 50,00 | 100,00 | 73,04 | 65,59 |
| Recall | 7,01 | 0,96 | 70,13 | 84,58 |
| ROC-AUC Score | 68,17 | 63,29 | 82,17 | 63,95 |

Based on the baseline models and a quick and dirty implementation, learning curves have shown that error is high for both instances: training and cross-validation sets regarding to:

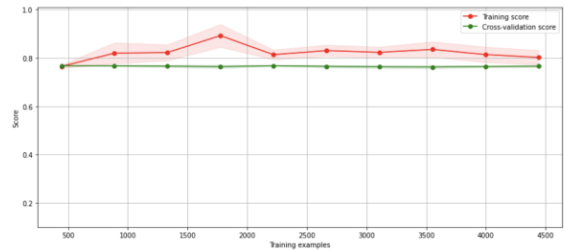

Fig. 10. Introversion and Extraversion learning curves

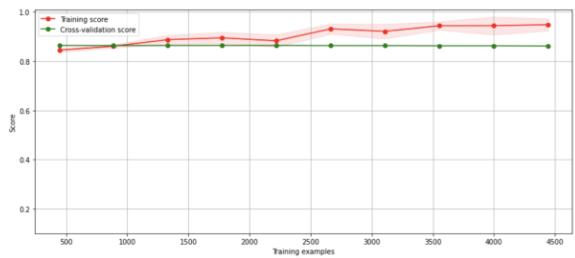

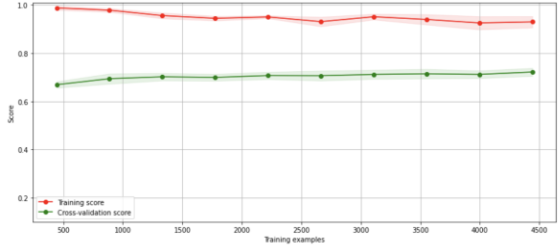

Fig. 11. Intuition and Sensing learning curves

Fig. 12. Thinking and Feeling learning curves

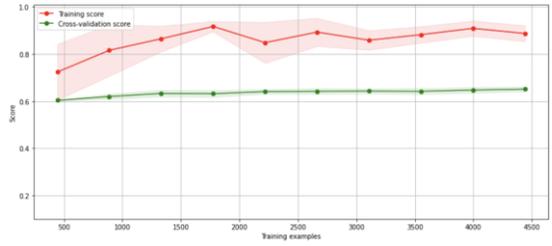

Fig. 13. Judging and Perceiving learning curves

With a little exception of J/P model, all of them are pretty underfitted because many mistakes are happening independently which curves is observed. Also, overfitting may occur for model's T/F and J/P because distance between lines is very high. Furthermore, confusion matrix[12] confirmed the diagnostic and reported the mistakes and successes that the stacked models had for each classifier. It has happened to all models under different proportions.

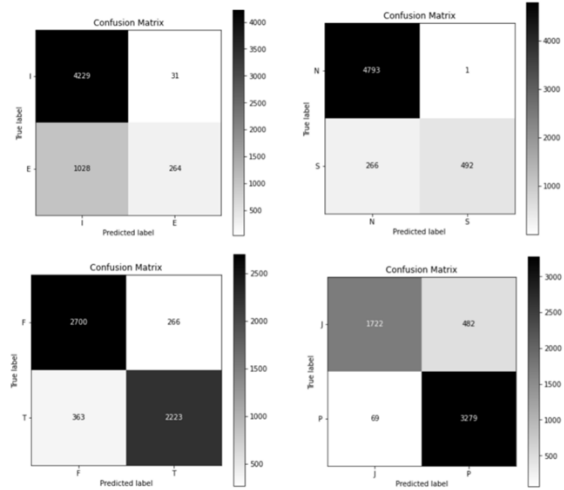

Fig. 14. Confusion Matrices for classifiers

### 4.3. Error & Ablative Analysis

Since the task proved to be complex in the initial rounds, there was a need to focus on the components of the pipeline to understand the benefit of any changes and also to foster work optimization. In the figure below it is possible to view the error and ablative analysis strategy.

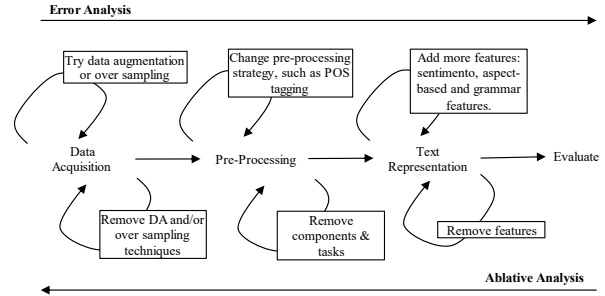

Fig. 15. Error & Ablative Analysis strategy

The top arrow represents the error analysis of the pipeline. This task started by verifying the balance of the training set and the possible measures to enrich the score of this component "Data Acquisition". Next, the techniques used for data pre-processing were evaluated, involving lemmatization, stopwords removal etc. From the evaluation of errors, the addition/removal of features was the most complex task since it was responsible for showing the possibilities of data articulation by improving the prediction and expanding the abstraction of personality types using MBTI. In it, the importance analysis of the features in relation to the model error were observed by paying attention to the next steps and for improving the overall score of the intelligent system. Finally, the use of more consolidated and/or complex models served as a comparison for the results obtained only with a focus on the data.

The other format for proving the hypotheses was the ablative analysis which, from the removal of some components, make (not only the specific importance of a certain component in the pipeline but) the importance of focusing on data (data-centric approach) in contrast standard optimizations to improve model performance (model-centric approach) clearer.

### 4.4. Experimental Protocol

An extremely important step for the execution of the strategy adopted in this study was the base pipeline for this. It was created in a robust way and already very optimized to remove possible influences during hyperparameter experiments, model assumptions, complexity, among others. Also, to avoid problems of data sanity, such as biased words, non-relevant texts etc.

---

[12] Also known as an error matrix, it is a specific table layout that allows visualization of the performance of an algorithm.

**Algorithm 2:** Logic Steps for Experimental Protocol

Description of the strategy used for designing, along the experiments, the text representation. In other words, how were the steps performed on data to achieve a good text representation for the study.

**REQUIRES:** dichotomy is comprised of four classifiers I+E, N+S, F+T and J+P. The experiments ran for each of them.

**FOR** each dichotomy **DO**:

    **LOOP ITERATION:**
        Data-Centric Approach: Data Iteration Loop (check Algorithm 1.)
    **ENDLOOP**

    **EXP. 1**
        **REQUIRES:** sentiment analysis and corresponding feature engineering.
        **RUN:** stacked model
        **FOR** each feature **DO:**
            Feature Importance
            Evaluation Metrics
        **ENDFOR**

    **EXP. 2**
        **REQUIRES:** aspect analysis and corresponding feature engineering.
        **RUN:** stacked model
        **FOR** each feature **DO:**
            Feature Importance
            Evaluation Metrics
        **ENDFOR**

    **EXP. 3**
        **REQUIRES:** grammar analysis and corresponding feature engineering.
        **RUN:** stacked model
        **FOR** each feature **DO:**
            Feature Importance
            Evaluation Metrics
        **ENDFOR**

    **EXP. 4**
        Put all together.
        **RUN:** stacked model
        **FOR** each feature **DO:**
            Feature Importance
            Evaluation Metrics
        **ENDFOR**

**ENDFOR**

In general, the assumptions defined in the first section of this document portray the benefit of improving data for both complex phenomenon and model performance. Here, data was benefit from a predominant focus through the pipelines proposed for the project and was observed throughout the experiment. Simplistically, "Algorithm 2" shows the steps needed in NLP for text classification task. This flow describes the experiments performed to optimize results from data-centric approach and to guarantee robustness.

In part, the idea of ensuring robustness to the model represents the current problem in generalizing broad tasks, such as: face detection (which can work with data from a camera, but not work with another type and so on). It is known that the current firmness and robustness of models is well linked to specific approaches. Thus, by analogy, it can be conceived that the experiment being conducted served to a specific purpose of being trained based on the personality types of the 8675 people who had 50 posts collected in the PersonaliteCafe forum and thereby inferring the types of related personalities.

### 4.5. Advanced Analytics

Previously the analyzes described below, the data set was pre-processed and its balance guaranteed (out of respect for the disproportionality between the types of personality available in the dataset which would get biased analyzes as result). The distribution was around 11 and 13% for each type of personality.

#### 4.5.1. Sentiment Analysis

According to the literature review (Lerman et al., 2009), (Silva and Perez, 2021), (Dowlagar and Mamidi, 2021) and the insights obtained during the study of the phenomenon, most psychologists attribute the reading of feelings to a relevant role for the study of personality types. Thus, a sentimental analysis was performed in an attempt to capture any additional patterns in the texts, their results were evaluated and new features were generated.

The extraction of feelings involved a complex linguistic analysis and detection of pattern matching which included the processing of grammatical classes, syntactic patterns, negation and so on, in order for the extraction to be carried out. As a result, the identified patterns and information about each extraction, including the type of information extracted (sentiment or request) were abstracted and analyzed. Sentiment rules were designed to extract information about someone's feelings about something. The rules extracted patterns that expressed the individual's feelings about concepts, places, actions, items and so on; and also categorize the feelings extracted in the following types of postures: (i) strong positive feeling, such as "great" or "excellent" (ii) weak positive feeling, such as "good" or "good", (iii) neutral feeling, such as "OK" or "acceptable", (iv) weak negative feeling, such as "bad" or "I don't like it", (v) strong negative feeling, such as "hate" or "terrible", (vi) minor problem, such as "useless" or "defective" and (vii) main problem, such as "broke" or "not working".

In addition, emoticons and a pejorative vocabulary set were used (profanity dictionary) segregated into ambiguous - which depend on the context - and unambiguous - which are always pejorative.

*Table 2*. Percentage of occurrences of the personality type in relation to all types for sentimental analysis.

| Sentiment (%) | I | E | N | S |
|---|---|---|---|---|
| General Request | 18,15 | 6,85 | 19,61 | 5,39 |
| Major Problem | 18,19 | 6,81 | 20,09 | 4,91 |
| Minor Problem | 18,42 | 6,58 | 19,60 | 5,40 |
| Neutral Sentiment | 18,46 | 6,54 | 19,54 | 5,46 |
| Request | 18,18 | 6,82 | 19,63 | 5,37 |
| Sentiment | 18,20 | 6,80 | 19,50 | 5,50 |
| Strong Negative | 18,32 | 6,68 | 19,46 | 5,54 |
| Strong Positive | 17,84 | 7,16 | 19,44 | 5,56 |
| Weak Negative | 18,46 | 6,54 | 19,41 | 5,59 |
| Weak Positive | 18,15 | 6,85 | 19,61 | 5,39 |

| Sentiment (%) | T | F | J | P |
|---|---|---|---|---|
| General Request | 13,01 | 11,99 | 10,99 | 14,01 |
| Major Problem | 12,41 | 12,59 | 10,33 | 14,67 |
| Minor Problem | 12,62 | 12,38 | 10,86 | 14,14 |
| Neutral Sentiment | 12,44 | 12,56 | 10,65 | 14,35 |
| Request | 13,03 | 11,97 | 11,01 | 13,99 |
| Sentiment | 12,37 | 12,63 | 10,54 | 14,46 |
| Strong Negative | 12,88 | 12,12 | 9,88 | 15,12 |
| Strong Positive | 11,36 | 13,64 | 10,43 | 14,57 |
| Weak Negative | 12,67 | 12,33 | 10,45 | 14,55 |
| Weak Positive | 12,64 | 12,36 | 10,72 | 14,28 |

From the sentimental portrait by personality types, it was possible to conclude some interesting relationships between demonstration and non-demonstration of feelings during writing. I and N show relatively more feelings in their writing than all the others, although P and F also do it with more distinction. Incredibly, extroverted people who are part of the sample tend not to show their feelings too strongly.

When a lot of time is spent in the same house, the human being tends to pay more attention to it; in the case of the introverted or intuitive individual, this happens in a more critical and constant manner. In the same way, it would happen with a car or with yourself. This analogy may explain why I and N have more general requests. On the other hand, the S tends not to express feelings; perhaps because of its practical and immediate side. "What is the use of proposing your feelings?" would be a way of thinking for them.

### 4.5.2. Aspect Analysis

In the analysis of aspects, the abstraction of the most critical aspects for the data set was evaluated. It was clarified that more than 35% of the identified aspects come from commercial content and, subsequently, geography presented great relevance in general (this applies to the four (4) dichotomies). It was possible to conceive that the expression of where the individuals stand, whether professionally or personally, matters for the abstraction of their personality types. Topics such as religion, medicine and sports were not so significant for the understanding of the individual's personality, not exceeding 1% of representativeness in the complete data set.

*Table 3*. Percentage of occurrences of the personality type in relation to all types for aspect analysis.

| Aspects (%) | I | E | N | S |
|---|---|---|---|---|
| Commercial | 18,25 | 6,75 | 18,97 | 6,03 |
| Country | 18,23 | 6,77 | 20,34 | 4,66 |
| Region Major | 21,88 | 8,61 | 2,44 | 6,06 |
| Educational | 19,42 | 5,58 | 18,40 | 6,60 |
| Media | 17,21 | 7,79 | 19,26 | 5,74 |
| Other | 18,16 | 6,84 | 21,28 | 3,72 |
| Entertainment | 18,66 | 6,34 | 18,25 | 6,75 |
| Government | 19,15 | 5,85 | 18,43 | 6,57 |
| Domestic | 18,65 | 6,35 | 18,94 | 6,06 |
| Water | 19,11 | 5,89 | 20,16 | 4,84 |

| Aspects (%) | T | F | J | P |
|---|---|---|---|---|
| Commercial | 13,19 | 11,81 | 10,37 | 14,63 |
| Country | 12,19 | 12,81 | 9,97 | 15,03 |
| Region Major | 15,71 | 14,78 | 12,52 | 17,98 |
| Educational | 11,67 | 13,33 | 10,58 | 14,42 |
| Media | 13,93 | 11,07 | 10,15 | 14,85 |
| Other | 13,13 | 11,87 | 7,70 | 17,30 |
| Entertainment | 11,96 | 13,04 | 10,53 | 14,47 |
| Government | 15,46 | 9,54 | 11,62 | 13,38 |
| Domestic | 11,63 | 13,37 | 7,40 | 17,60 |
| Water | 13,61 | 11,39 | 11,91 | 13,09 |

Additionally, it can be mentioned that aspects such as entertainment and media had a good representation, around 9%.

### 4.5.3. Grammar Analysis

*Table 4*. Percentage of occurrences of the personality type in relation to all types for grammar analysis.

| Syntactic Terms (%) | I | E | N | S |
|---|---|---|---|---|
| Main Verb - Active | 20,80 | 7,91 | 22,36 | 6,34 |
| Direct Object | 20,43 | 7,71 | 21,92 | 6,22 |
| Subject | 23,32 | 8,94 | 2,51 | 0,71 |
| Predicate | 18,97 | 7,31 | 20,58 | 0,57 |
| Main Verb - Copula | 2,45 | 9,20 | 26,65 | 7,04 |
| Predicate Subject | 21,90 | 8,45 | 2,38 | 6,59 |
| Indirect Object | 18,20 | 6,80 | 19,45 | 5,55 |
| MainVerb - Passive | 16,89 | 8,11 | 20,95 | 4,05 |

| Syntactic Terms (%) | T | F | J | P |
|---|---|---|---|---|
| Main Verb - Active | 14,25 | 14,45 | 12,24 | 1,65 |
| Direct Object | 14,20 | 1,39 | 11,90 | 16,24 |
| Subject | 16,07 | 16,19 | 13,71 | 18,55 |
| Predicate | 12,65 | 13,63 | 10,99 | 15,30 |
| Main Verb - Copula | 16,56 | 17,13 | 1,41 | 19,57 |
| Predicate Subject | 14,65 | 15,70 | 12,69 | 17,66 |
| Indirect Object | 12,74 | 12,26 | 10,71 | 14,29 |
| MainVerb - Passive | 12,16 | 12,84 | 9,46 | 15,54 |

People P and I tend to use the passive voice more - a verbal voice that indicates that the subject of the speaker suffers or receives a certain action, instead of practicing it. This insight for P seems natural since this personality trait prefers to react to the environment rather than trying to control it while the I is sensitive to external stimuli and therefore prefers passive language.

Copulas (mainly is and are), in a way, "define" things that the J profile tends to keep apart due to its strong work ethic, placing its duties and responsibilities above all. Thus, it is not common to see them defining abstract things, such as feelings, ideas, etc. An Italian corpus study (Bassignana et al., 2020) evaluated Italian words against personality types and the result illustrates that the grammatical abstraction in reference to personality does not depend on the nationality or characteristics of the language, but rather on the individual's form of expression.

---

[13] Factorization of a matrix that generalizes the eigendecomposition matrix to any *m* x *n* matrix via an extension of the polar decomposition

### 4.5.4. Feature Importance

Feature importance analysis was conducted to better clarify what exactly in the features is or is not robust for each experiment. Principal Component Analysis (PCA) (Zhang, 2019) was used to determine feature importance. Linear dimensionality reduction was achieved by using Singular Value Decomposition (SVD)[13] of the data to project it to a lower dimensional space which the input data is centered but not scaled for each feature before SVD application. The number of components to keep was obtained with the formula n_components == min(n_samples, n_features) – 1 and the solver was selected by a default policy based on shape of X and n_components.

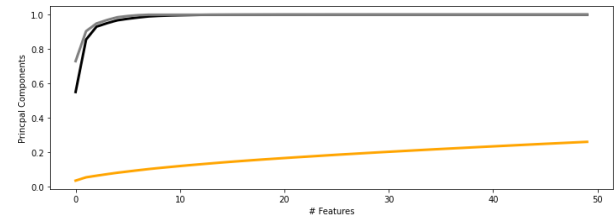

Fig. 16. Percentage of variance explained by each of the selected components (all components are stored and the sum of the ratios is equal to 1.0).

According to the graph above, original features could explain only around 15% of the variance in the dataset with the first 10 principal components which is not good while sentiment, grammar and aspect features could explain much faster, suggesting their importance is superior to the baselines. Moreover, the correlations between baseline features and principal components could only provide poor correlations, and this confirmed once again underfitting diagnosis. With the addition of new features based on technical and domain knowledge, it can be seen that there was a significant evolution in how important the features have become more important for the models through their correlation with the main components.

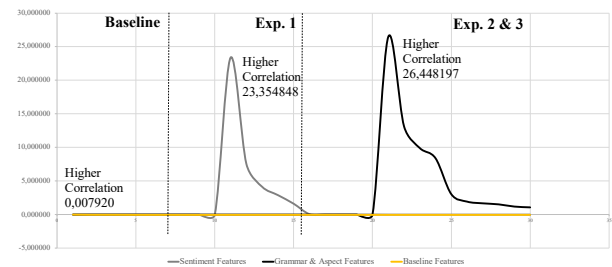

Fig. 17. Correlation by Experiments and its increasing (%).

As a result, features that most contributed under PCA results had been inserted into the model. It turned out that a reasonable improvement was noticed in evaluation metrics for the models; however, in part, this was due to the lack of explanatory power of the other features or their irrelevance. Going deeper into the analysis of these variables that most contribute to the explanation of the phenomenon to the model, we have:

*Table 5*. Top 10 words which most contributed for model results.

| Top 10 Features | Importance (%) |
|---|---|
| mainverb/active | 26,45 |
| Sentiment | 23,35 |
| subject | 13,14 |
| directobject | 9,90 |
| Sentiment | 8,38 |
| WeakPositiveSentiment | 7,56 |
| StrongPositiveSentiment | 4,15 |
| WeakPositiveSentiment | 3,03 |
| MinorProblem | 2,90 |
| StrongPositiveSentiment | 1,93 |

It can be conceived that even though these features helped to increase the feature relevance to the model, the contribution must be understood. For example, the "subject" feature will always have a good expression in the dataset since its representativeness is high and, therefore, it cannot be explained whether or not it is relevant. However, "Weak Positive Sentiment" may mean that, due to its presence being more significant in the observations of I and N, the demonstration of weak and positive feelings is important to identify these respective types of personalities, etc.

### 4.6. Final Results

The tables below show the comparison between the models that comprised all experiments for data-centric tasks (feeling, aspects, grammar and everything) and more complex models (greater implementation effort), such as LSTM and BERT Pre-Trained Model. As provided in the previous sections, the experiments consisted of: (i) sentimental analysis and creation of new variables for inclusion in the stacked model, (ii) aspect-based sentimental analysis to project the most relevant aspects for the prediction of personality types; also, creation of variables, (iii) grammatical analysis by focusing on the writing and technical expression of individuals as well as the generation and inclusion of variables in the stacked model and (iv) the joining of all variables was carried out by generalizing specific benefits for each dichotomy.

Finally, a sequential model of deep and pre-trained neural networks (representing the state of the art) were used in contrast to the improvement of metrics and speed of implementation (still considering possible costs with implementations or couplings in productive systems).

*Table 6*. Metrics comparison for Intraversion x Extraversion (I/E) model. In bold, the best metrics obtained through the experiments which showed that sentiment analysis would achieve better results on model and prediction power quality than Bert, a state-of-art model for text classification task.

| Metrics | Baseline | Sentiment | Aspects | Grammar | All | LSTM | BERT |
|---|---|---|---|---|---|---|---|
| Accuracy | 76,37% | **78,96%** | 76,73% | 76,59% | 78,60% | 76,69% | 76,46% |
| F1-Score Macro | 49,32% | 58,39% | 49,29% | 48,74% | 57,25% | 43,40% | **59,55%** |
| F1-Score Micro | 76,37% | **78,96%** | 76,73% | 76,59% | 78,60% | 76,69% | 76,46% |
| F1-Score Weighted | 68,85% | 73,82% | 68,96% | 68,66% | 73,18% | 66,58% | **73,84%** |
| F1-Score | 12,00% | 29,00% | 12,00% | 11,00% | 27,00% | 0,00% | **33,00%** |
| Jaccard Score | 6,55% | 17,05% | 6,38% | 5,80% | 15,62% | 0,00% | **20,04%** |
| Log Loss | 8,16% | **7,27%** | 8,04% | 8,09% | 7,39% | 8,05% | 8,13% |
| Precision | 50,00% | **71,43%** | 56,41% | 54,05% | 69,62% | 0,00% | 46,60% |
| Recall | 7,01% | **18,29%** | 6,71% | 6,10% | 16,77% | 0,00% | 26,02% |
| ROC-AUC Score | 68,17% | **73,92%** | 69,01% | 68,75% | 71,61% | 50,18% | 69,74% |

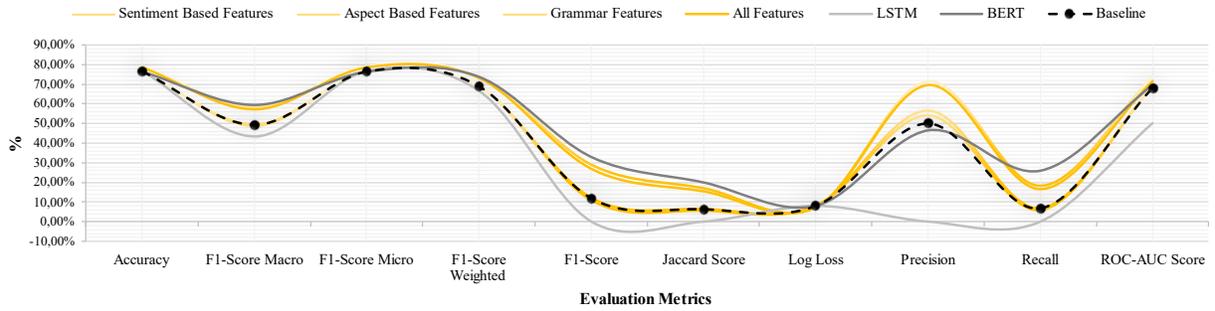

Fig. 18. Evolution of evaluation metrics against experiments for Intraversion x Extraversion (I/E) model

In general, the evolution of the model's evaluation metrics due to the enrichment of variables proved to be more satisfactory under the use of sentiment analysis - which was already expected according to the studied literature and benchmarks; however, the features derived from the grammatical and aspect analysis, although it had a good importance in the PCA analysis, did not show promise for the improvement of the model. The "All" column refers to the mixture of all features and for the identification of I/E personality types, the sum was not beneficial. From the psychological perspective, we can conceive that the introspective individual usually finds strength in the creation of his own space that can be textual while the extrovert is interested in involvement with the environment (even if virtual). Although introspective individuals have a greater presence in the dataset, E has approximately only 30% in relation to it, both have a common bias for textual expression.

*Table 7*. Metrics comparison for Intuition x Sensing (N/S) Model. Bold metrics show that sentiment and grammar are better contributors for performance than others for model quality perspective; however, for distinguishing capacity are most due to grammar features. It looks like abstracting text grammar contributes best for improving distinguishing power for the model.

| Metrics | Baseline | Sentiment | Aspects | Grammar | All | LSTM | BERT |
|---|---|---|---|---|---|---|---|
| Accuracy | 85,16% | 85,95% | 85,09% | 85,16% | 85,73% | 85,41% | **86,36%** |
| F1-Score Macro | 46,94% | **53,60%** | 46,91% | 46,94% | 52,69% | 46,07% | 46,34% |
| F1-Score Micro | 85,16% | 85,95% | 85,09% | 85,16% | 85,73% | 85,41% | **86,36%** |
| F1-Score Weighted | 78,47% | **80,73%** | 78,44% | 78,47% | 80,38% | 78,69% | 80,03% |
| F1-Score | 2%% | 15,00% | 2,00% | **20,00%** | 13,00% | 0,00% | 0,00% |
| Jaccard Score | 0,96% | **8,02%** | 0,96% | 0,96% | 7,04% | 0,00% | 0,00% |
| Log Loss | 5,13% | 4,85% | 5,15% | 5,13% | 4,93% | 5,04% | **4,71%** |
| Precision | 100,00% | 80,95% | 66,67% | **100,00%** | 75,00% | 0,00% | 0,00% |
| Recall | 0,96% | **8,17%** | 0,96% | 0,96% | 7,21% | 0,00% | 0,00% |
| ROC-AUC Score | 63,29% | 66,73% | 62,63% | 62,82% | **65,88%** | 49,89% | 53,96% |

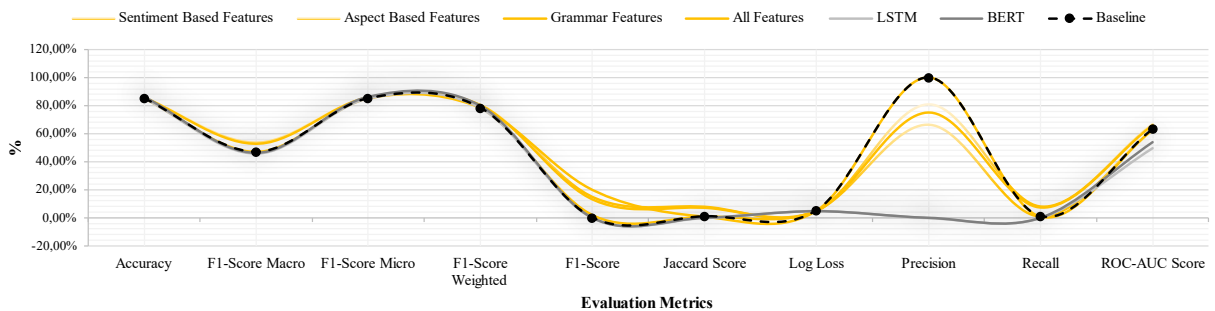

Fig. 19. Evolution of evaluation metrics against experiments for Intuition x Sensing (N/S) model

Intuitive individuals (N) or observers (S) work relatively close to grammar, which is the resource that allow them, in writing, to imagine the potential of past and future tenses. However, S is more interested in facts and more direct results. In general, they like to see the "Big Picture" and love theory, which can give them a better grammatical structure. Still, it could be observed that the variables of feelings helped and corroborated for a better understanding of these personality types, since the sum of grammar and feelings allows a better conception of the meaning invested in the text. Obviously, it is known that many current models already deal with irony or sarcasm detection, for example; however, initial advances in the data-centric approach already anticipate benefits similar to the use of these new neural network architectures. Lastly, as dataset is unbalanced, N has a representativeness of approximately 16% in relation to S.

*Table 8*. Metrics comparison for Feeling x Thinking (F/T) Model. As expected, this dichotomy predominantly benefits from sentimental analysis and abstraction and, therefore, it was possible to conceive faster and greater results than with the use of more complex models or hyperparameters tuning.

| Metrics | Baseline | Sentiment | Aspects | Grammar | All | LSTM | BERT |
|---|---|---|---|---|---|---|---|
| Accuracy | 74,86% | **75,43%** | 74,57% | 73,70% | 74,35% | 48,97% | 68,29% |
| F1-Score Macro | 74,51% | **75,15%** | 74,22% | 73,33% | 73,94% | 32,87% | 67,71% |
| F1-Score Micro | 74,86% | **75,43%** | 74,57% | 73,70% | 74,35% | 48,97% | 68,29% |
| F1-Score Weighted | 74,80% | **75,41%** | 74,51% | 73,64% | 74,26% | 32,20% | 67,44% |
| F1-Score | **72,00%** | **72,00%** | 71,00% | 70,00% | 71,00% | 66,00% | **72,00%** |
| Jaccard Score | 55,71% | 56,84% | 55,32% | 54,03% | 54,65% | 48,97% | 56,27% |
| Log Loss | 8,68% | **8,49%** | 8,78% | 9,08% | 8,86% | 17,63% | 10,95% |
| Precision | 73,04% | 73,25% | 72,71% | 71,86% | 72,96% | 48,97% | 61,37% |
| Recall | 70,13% | 71,73% | 69,81% | 68,53% | 68,53% | **100,00%** | 87,14% |
| ROC-AUC Score | 82,17% | 82,89% | 81,82% | 81,13% | 81,93% | 49,94% | 79,20% |

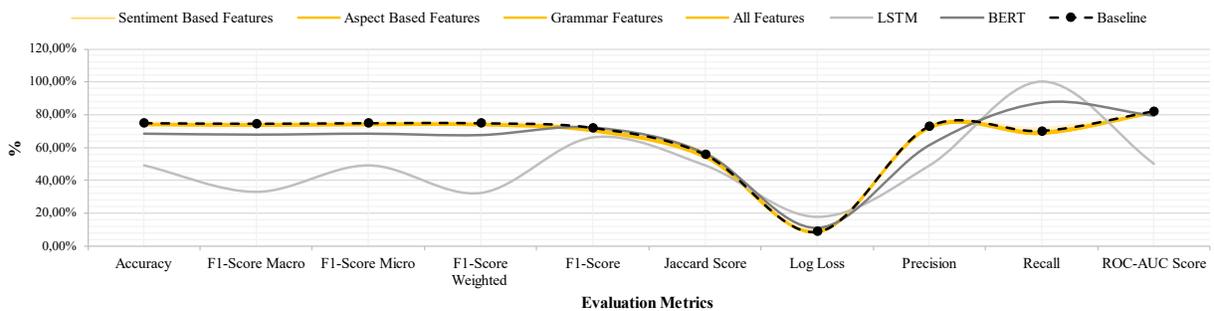

Fig. 20. Evolution of evaluation metrics against experiments for Feeling x Thinking (F/T) model

Thinking or T individuals focus on objectivity and rationality, prioritizing logic over emotions. They tend to hide their feelings while Fs are sensitive and emotionally expressive. Although the dataset was relatively balanced for this dichotomy and there was no bias that denoted the representation of feelings for the T class, it was observed that "hiding the feelings" was not very adherent to the findings. Of course, there are studies that suggest that on the internet the virtual personalization of an individual is extremely common and because of that they behave as if they were someone else. Perhaps this explains why a personality type so retracted in the expression of its feelings is exposing it so much in the posts.

*Table 9*. Metrics comparison for Judging x Perceiving (J/P) Model. This was a clear example where experiments in favor of psychological abstraction of data through textual analysis and handcrafted features could not collaborate significantly in contrast to more complex models.

| Metrics | Baseline | Sentiment | Aspects | Grammar | All | LSTM | BERT |
|---|---|---|---|---|---|---|---|
| Accuracy | 63,69% | 63,33% | 63,47% | 62,18% | 64,19% | 60,63% | **64,29%** |
| F1-Score Macro | 57,16% | 56,59% | 57,27% | 55,34% | 57,72% | 37,74% | **61,64%** |
| F1-Score Micro | 63,69% | 63,33% | 63,47% | 62,18% | 64,19% | 60,63% | **64,29%** |
| F1-Score Weighted | 60,75% | 60,26% | 60,76% | 59,09% | 61,27% | 45,77% | **63,84%** |
| F1-Score | 74,00% | 74,00% | 74,00% | 73,00% | 74,00% | **75,00%** | 72,00% |
| Jaccard Score | 58,59% | 58,35% | 58,17% | 57,25% | 59,06% | **60,63%** | 55,92% |
| Log Loss | 12,54% | 12,67% | 12,62% | 13,06% | 12,37% | 13,60% | **12,33%** |
| Precision | 65,59% | 65,29% | 65,64% | 64,61% | 65,90% | 60,63% | **69,27%** |
| Recall | 84,58% | 84,58% | 83,63% | 83,39% | 85,05% | **100,00%** | 74,37% |
| ROC-AUC Score | 63,95% | **64,91%** | 63,59% | 64,00% | 64,46% | 49,96% | 63,73% |

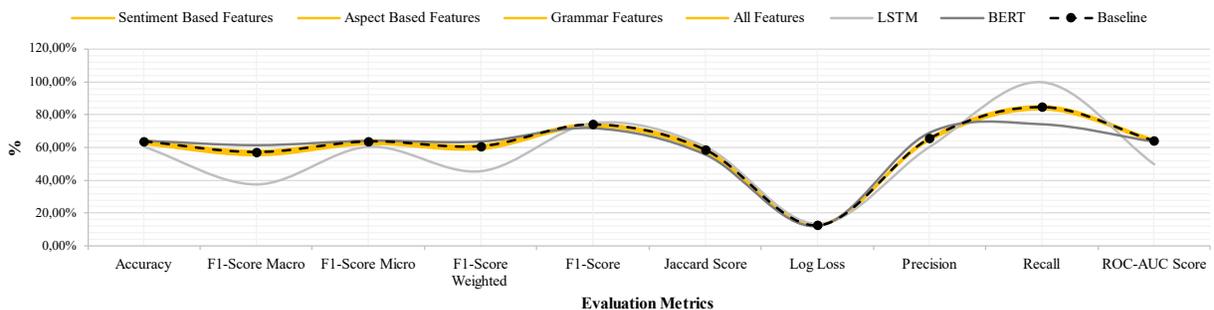

Fig. 21. Evolution of evaluation metrics against experiments for Judging x Perceiving (J/P) model

Judging individuals J are decisive and, generally, organized and not so spontaneous; therefore, they may feel uncomfortable expressing their ideas through dynamic resources (a forum, for example) while P individuals have high flexibility to take advantage of unexpected opportunities and end up reacting to their environments instead of trying to control it, which may lead to non-commitment to something uncertain or doubtful (like giving your opinion on a forum, for example). The personality type J appears in the dataset 35% more than the P and this can lead to the lack of expression needed to read the personality type in the posts. This may explain why the extremely complex models with an impressively large number of trainable parameters did not do so well or why the JP models underperformed the other models.

## 5. Discussion

### 5.1. Significance & Proof

As described, the task of classifying personality types is by far one of the most difficult in the context of psychological types, since aspects with greater relevance are not easily obtained (geography, society, culture, storm, etc.). For this reason, it can affirm the relevance of the study, exclusively, for the trained scenario - individuals who posted texts on the Personalite Café forum to discuss their personality traits. However, the study conducted showed qualitatively and quantitatively that the benefit of the data-centric approach to the overall performance of the experiment can be extended to other applications or scenarios. Through the domain of knowledge (psychology and personality types), it was possible to design analyzes more directed to the data: sentimental, grammatical and aspects.

These analyzes were fundamental to exercise domain knowledge in the representation of texts in NLP (or resource engineering) and, as a result, handcrafted features were created in the hope of optimizing the abstraction of entities and patterns in the dataset. The functional hypothesis proved to be true since these characteristics, in fact, provided a significant increase in the importance, quality and explanatory power performance for the models used in the study.

Obviously, the generation of features with consistent rationale obtained from the study of the area/domain established a better connection and representation of the dataset phenomenon than features generated from the frequency of terms or other techniques that were used for the representation of texts. More complex models used more specific tokenization methods and although they had provided a good text representation, it could not observe meaningful superiority of their performances in the experiments against the simplest.

From the technical point of view, it can be conceived that the hypothesis that data-centric models showed improvement in numbers for the first experiments and with less time than with the use of more complex models (LSTM took on average between 20 per training by dichotomy while the BERT on average 2h30). In other words, the data-driven approach really brought performance improvements faster resulting on the second hypothesis confirmation.

### 5.2. Findings Implications and Limitations

Throughout the study, it was possible to summarize the three most relevant implications or limitations for the sequence, the enrichment of the approach that are described below.

**Unbalanced Dataset**

Some implications were due to the imbalance of the data; so techniques such as Data Augmentation and oversampling may prove to be efficient for improving this item and can be worked on any of the four (4) classifiers.

**Biased Dataset**

Since it already had more information about the personality types that are most likely to themselves online. For this aspect, the acquisition of new data (update of personalite cafe data) and/or insertion of data from other blogs, forums can significantly help in the abstraction of other topics. In particular, the JP model can benefit more from new data since the current observations have not proven to be very significant for detecting this dichotomy in the presented results.

**Generalization**

Even if some narratives (Ramezani et al., 2021) suggests that human personality is significantly represented by words and also how technology is becoming more human (Clayton, 2014) or provide to machines the ability to have engaging conversation, with personality and empathy (Shuster et al., 2020), the generalization of this study implies a broad spectrum that was not fully conceived and is therefore recommended in new samples from the personalite cafe forum or similar forums since generalization to other scenarios requires extremely important features that were not observed or focused on in these experiments.

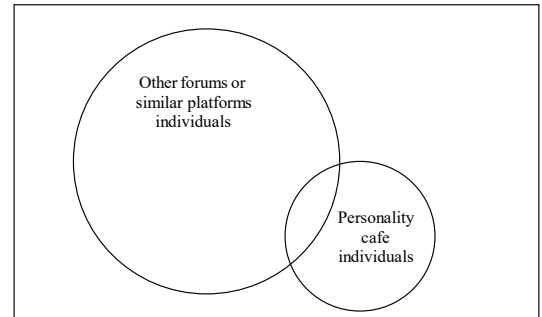

Fig. 22. Promising generalization sets

In addition, from the approach taken, the design of new formats, pipelines, extensions and new models can provide a significant room for improvements. This is because the focal point of this study was not established in a comprehensive perspective, but in a specific one for justifying the hypotheses raised. Therefore, other implications or limitations may exist through new approaches that were not detected in this work.

### 5.3. Directions and Future Research

As a result of this work, some directions for future research were detected and described below in order to streamline the design of new studies and the identification of opportunities for the observed phenomenon. It was possible to conceive that, from the complexity of identifying personality types, many perspectives for future research directions are very likely to be expanded. Impressive results may come along with different approaches and this work really expects to be a little help towards the extension of personality types abstraction. By the way, the following topics can have even more potential if data-centric approach is kept.

**Domain knowledge**

Domain knowledge expansion is expected to give further abilities for studying other psychological features for better performance of the conducted experiments, such as, for example, signs of character, temperament, other classification methods other than MBTI, such as Keirsey Temperament Sorter[14], Socionics[15], Big-Five[16], Big-Two among other techniques.

---

[14] Keirsey official page.
[15] Socionics web site for further information.

[16] Hierarchical structure of the Big Five

**Systematic Data-Centric Approach**

Methods for implement data-centric model by using a systematic way through innovative tooling are promising to achieve even better ways of improving data quality for the models and go live with machine learning system into a productive environment.

**Text Representation (Feature Engineering)**

The approach of new analyzes for the generation of handcrafted features could establish new expanding levels for the representation of the phenomenon, such as, for example, other types of sentimental analysis, graph analysis, other tokenization methods, data augmentation ([Liesting et al., 2021](#)), etc. Use of other methods for assessing importance ([Lee et al., 2021](#)) and data quality ([Azeroual and Abuosba, 2017](#)), including psychological.

**Other Data Sources**

Still, the expansion of the dataset used to consider other forums or similar sources can be a good extension of the work since the enrichment of the source obtained shows a promising performance improvement. The results of this study may serve as a "signal" within larger pipelines to extend the ability to identify personality types to other approaches, such as recommendation systems, prevention of suicidal tendencies, identification of psychic pathological signs, among many others.

## 6. Conclusions

Today, the reading of human personality is based on time-dependent variables (still) impossible to be generated. That is, without sufficient depth for abstraction of human expression throughout its history it is not possible to achieve high levels of real accuracy and psychological understanding.

However, the results of this study proved that by using a data-centric approach based on features resulting from analyzes supported by knowledge of the area/domain (psychology), it is possible to extend the abstraction of personality types significantly and quickly using Machine Learning (ML) and Natural Language Processing (NLP). Through sentimental, grammatical and aspects analysis, handcrafted features were generated and, together with the importance evaluation of these features, it supplied a rapid and relevant improvement in the models' performance against complex and state of art models.

The use of a robust baseline that had a stack model (ensemble method) and optimized hyperparameters setting (grid search) for such an experiment ensured that the data iteration loop was possible in a simpler and more efficient way and thus, systematically, the proposal to work with the features in the generated pipelines was established quickly and easily.

It was expected that this work was a needle in a haystack because of the existence of many possibilities and possible applications for the prediction of personality types. However, it was possible to provide results with high significance - a very good needle - to current and future efforts for the task and establish a huge room for improvement on the field.

If we go a little further, a baby step in this direction, one day we will have the opportunity to develop something closer to the real and thus improve the world from the knowledge of ourselves.